\documentclass[10pt,twocolumn,letterpaper]{article}

\usepackage{cvpr}              
\usepackage{amsmath}
\usepackage{graphicx}
\usepackage{svg}
\usepackage{adjustbox}
\usepackage{caption}
\usepackage{subcaption}
\usepackage[figurename=Fig.]{caption}
\usepackage{tabularx} 

\definecolor{cvprblue}{rgb}{0.21,0.49,0.74}
\usepackage[pagebackref,breaklinks,colorlinks,citecolor=cvprblue]{hyperref}

\newcommand*\samethanks[1][\value{footnote}]{\footnotemark[#1]}

\def\paperID{22} 
\def\confName{CVPR}
\def\confYear{2024}
\newcommand{\bigO}{\mathcal{O}}

\begin{document}
\title{SegFormer3D: an Efficient Transformer for 3D Medical Image Segmentation}

\author{Shehan Perera\thanks{These authors contributed equally to this work} , Pouyan Navard\samethanks, Alper Yilmaz \\ 
Photogrammetric Computer Vision Lab, The Ohio State University \\
{\tt\small \{perera.27, boreshnavard.1, yilmaz.15\}@osu.edu}
}

\maketitle

\begin{abstract}
The adoption of Vision Transformers (ViTs) based architectures represents a significant advancement in 3D Medical Image (MI) segmentation, surpassing traditional Convolutional Neural Network (CNN) models by enhancing global contextual understanding. While this paradigm shift has significantly enhanced 3D segmentation performance, state-of-the-art  architectures require extremely large and complex architectures with large scale computing resources for training and deployment. Furthermore, in the context of limited datasets, often encountered in medical imaging, larger models can present hurdles in both model generalization and convergence. In response to these challenges and to demonstrate that lightweight models are a valuable area of research in 3D medical imaging, we present SegFormer3D, a hierarchical Transformer that calculates attention across multiscale volumetric features. Additionally, SegFormer3D avoids complex decoders and uses an all-MLP decoder to aggregate local and global attention features to produce highly accurate segmentation masks. The proposed memory efficient Transformer preserves the performance characteristics of a significantly larger model in a compact design. SegFormer3D democratizes deep learning for 3D medical image segmentation by offering a model with $33\times$ less parameters and a $13\times$ reduction in GFLOPS compared to the current state-of-the-art (SOTA). We benchmark SegFormer3D against the current SOTA models on three widely used datasets Synapse, BRaTs, and ACDC, achieving competitive results. Code: \href{https://github.com/OSUPCVLab/SegFormer3D.git}{https://github.com/OSUPCVLab/SegFormer3D.git}
\end{abstract}

\begin{figure}
    \centering
    \includegraphics[width=0.475\textwidth]{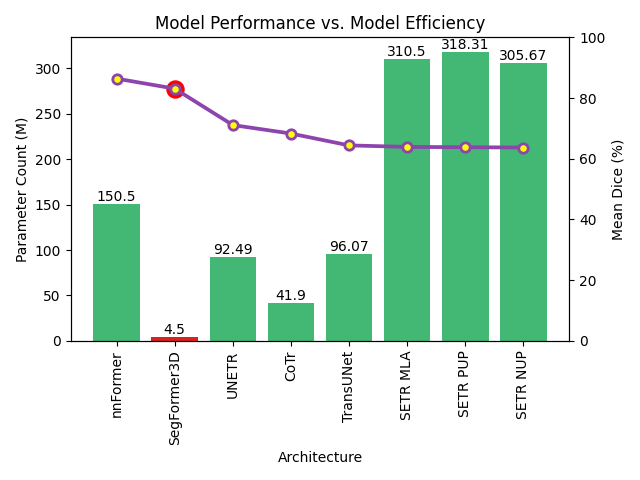}
    \captionsetup{font={scriptsize}}
    \vspace{-6pt}
    \caption{
    \textbf{Parameter Count \vs Performance on BraTs} We compare Segformer3D to existing 3D volumetric image segmentation architectures evaluating model performance with respect to parameter count.  
    The green bars represent model parameters while the purple plot shows the mean dice performance for each architecture. We demonstrate that at 4.5 million parameters Segformer3D is a highly competitive lightweight architecture for 3D medical image segmentation.}
    \label{fig:fig1}
    
    \vspace{-16pt}
\end{figure}

\section{Introduction}

The emergence of deep learning in healthcare has been transformative, offering an unprecedented capacity to learn and analyze complex medical data patterns. A fundamental task in medical image analysis is 3D volumetric image segmentation that is crucial for applications such as tumor and multi-organ localization in diagnosis and treatment. The conventional approach involves employing an encoder-decoder architecture \cite{ronneberger2015u,  long2015fully}, where the image is first transformed into a low-dimensional representation, and then the decoder maps the representation to a voxel-wise segmentation mask. However, these architectures struggle to generate accurate segmentation masks due to their limited receptive field. Recently, Transformer-based techniques have demonstrated superior segmentation performance owing to the ViT's ability to utilize attention layers for capturing global relationships~\cite{hatamizadeh2022unetr,zhou2021nnformer}. This stands in sheer contrast to CNNs which exhibits local inductive bias properties.

\begin{figure*}[ht]
    \centering
    \includegraphics[width=\textwidth, height=10cm]{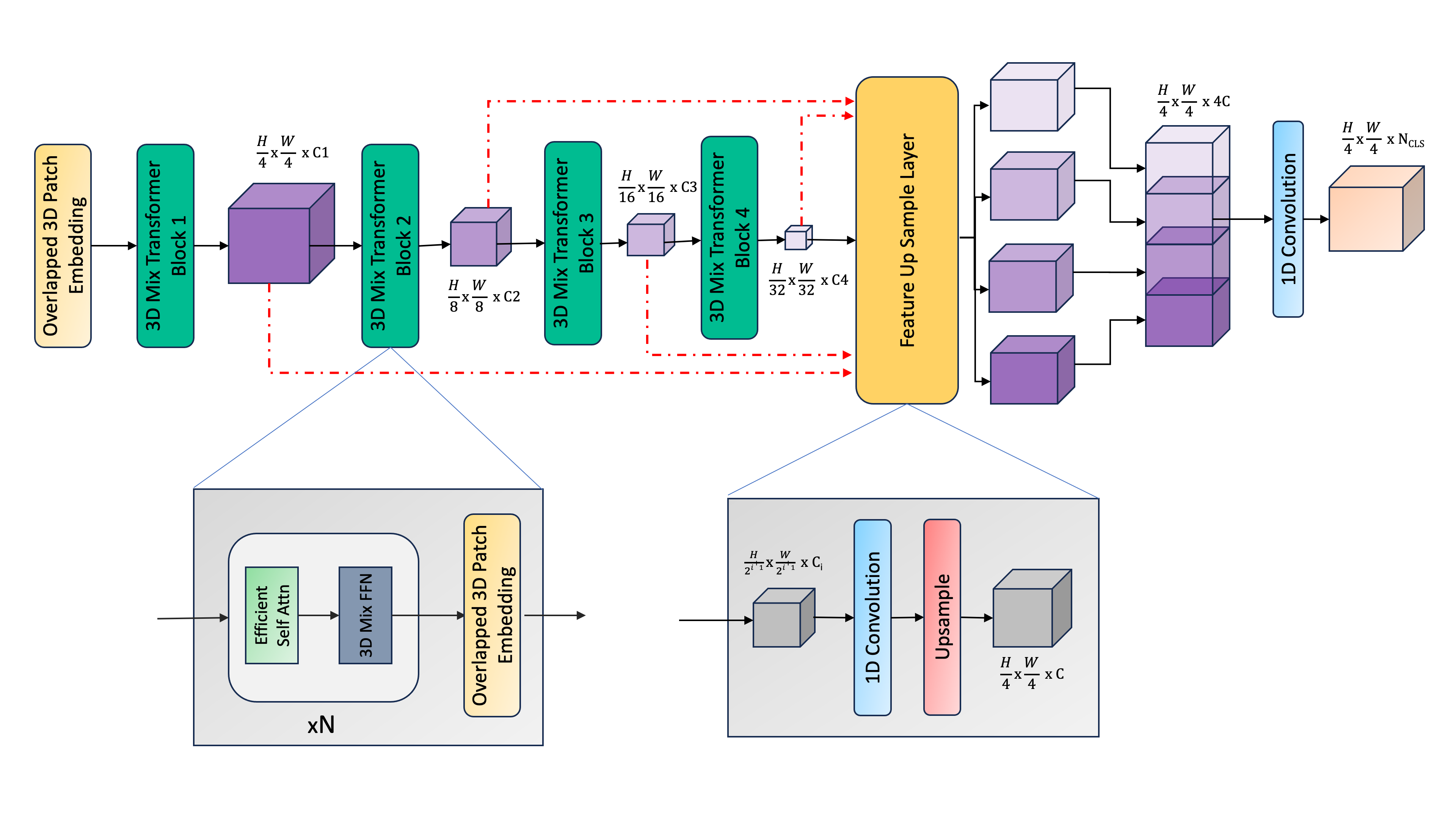} 
    \caption{Segformer3D Overview: The model input is a 3D volume $\mathbb{R}^{D \times C \times H \times W}$. We extract multiscale volumetric features using a 4 stage hierarchical Transformer. An all-MLP decoder then upsamples and aggregate local and global attention features from the encoding stage to generate the final segmentation mask.}
\end{figure*}

Following the seminal work of TransUnet\cite{chen2021transunet} and UNETR \cite{hatamizadeh2022unetr}, a large body of research in the medical community has been dedicated to designing Transformer based architectures that take advantage of the strong encoding capability of ViTs and the feature refinement capability of CNNs in the decoding stage. For example, \cite{hatamizadeh2021swin, hatamizadeh2022unetr, zhou2021nnformer} combined localized receptive field of convolutions and global attention. Despite their advantages, ViTs fail to match the generalization capabilities of CNNs when trained from scratch on small-scale datasets, and because of lack of inductive bias often depends on large scale datasets for pretraining \cite{dosovitskiy2020image} which are not commonly available in medical image domain. Additionally, computational efficiency of the ViTs are bounded by number of floating point operation and element wise functions in the multi head self attention block \cite{liu2023efficientvit}. This issue is much more prominent in 3D medical imaging tasks because of the fact that the length of the converted sequence of the 3D volumetric input is considerably long. Furthermore, medical imaging data frequently exhibits repetitive structures \cite{cciccek20163d}, suggesting that it can be compressed, a consideration often overlooked by the 3D SOTA ViT architectures in the medical domain.

This paper presents SegFormer3D, a volumetric hierarchical ViT, that extends \cite{xie2021segformer} to 3D medical image segmentation tasks. Unlike vanilla ViT\cite{dosovitskiy2020image} which renders feature maps on a fixed scale, Segformer3D encodes feature maps at different scales of the input volume following the Pyramid Vision Transformer \cite{wang2021pyramid}. Our design enables the Transformer to capture a variety of coarse to fine-grained features of the input. SegFormer3D also utilizes an efficient self-attention module \cite{wang2021pyramid} that compresses the embedded sequence to a fixed ratio to significantly reduce model complexity without sacrificing performance Figure \ref{fig:fig1}. Additionally, SegFormer3D utilizes the overlapping patch embedding module used in \cite{xie2021segformer} that preserves the local continuity of the input voxels. This embedding uses a positional-free encoding \cite{islam2020much} that prevents accuracy loss when there is a resolution mismatch during training and inference, which is common in medical image segmentation. To efficiently generate a high-quality segmentation mask, SegFormer3D uses an all-MLP decoder introduced in \cite{xie2021segformer}. Comprehensive experiments on three benchmark datasets- Synapse\cite{landman2015miccai}, ACDC\cite{bernard2018deep}, and BRaTs\cite{menze2014multimodal}—validate the qualitative and quantitative effectiveness of SegFormer3D. Our contributions can be summarized as:

\begin{itemize}
    \item We introduce a lightweight memory efficient segmentation model that preserves the performance characteristics of larger models for 3D medical imaging.
    \item With 4.5 million parameters and 17 GFLOPS, Segformer3D presents a $34\times$ and $13\times$ reduction in parameter count and model complexity vs SOTA. 
    \item We showcase highly competitive results without pretraining, emphasizing the generalization capabilities of lightweight ViTs and that exploring architectures like Segformer3D is a valuable research area in medical imaging.
\end{itemize}

\section{Related Work}

Following the introduction of Unet\cite{ronneberger2015u}, numerous approaches have been proposed for medical image analysis such as Dense-unet\cite{cai2020dense} and deep-supervised CNN \cite{zhu2017deeply}. Unet has also been extended to 3D medical image analysis, for instance, 3D-Unet\cite{cciccek20163d}, V-net\cite{milletari2016vnet}, nn-Unet\cite{isensee2018nnunet} and \cite{gibson2018automatic, dou20163d, roth2017hierarchical}. Researchers also designed hierarchical architectures to capture contextual information. In \cite{milletari2016vnet}, Milletari et al. downsampled the volume to a lower resolution to preserve beneficial image features using V-net. Cicek et al.\cite{cciccek20163d} replaced the 2D to 3D convolutions in 3D-unet. Isensee et al.\cite{isensee2018nnunet} proposed the nn-Unet generalized segmentation architecture that can extract features at multiple scales. In \cite{li2020pgdunet}, PGD-UNet uses deformable convolution to deal with irregular organ shapes and tumors for medical image segmentation.

Several recent papers have studied Transformer-convolution architectures such as TransUnet\cite{chen2021transunet}, Unetr\cite{hatamizadeh2022unetr},  SwinUnetr\cite{hatamizadeh2021swin}, ,TransFuse\cite{zhang2021transfuse}, nnFormer\cite{zhou2021nnformer} and LoGoNet\cite{monsefi2024masked}. TransUnet\cite{chen2021transunet} combines Transformers and U-Net to encode image patches and decode through high resolution upsampled CNN features for localization. Hatamizadeh et al. \cite{hatamizadeh2022unetr} present UNETR, a 3D model merging the long-range spatial dependencies characteristic of Transformers with the inherent CNN inductive biases in a "U-shaped" encoder-decoder structure. In UNETR, Transformer-blocks encode features that capture consistent global representations and are subsequently integrated across various resolutions within a CNN-based decoder. LoGoNet\cite{monsefi2024masked} uses Large Kernel Attention (LKA) and a dual encoding strategy to capture both long-range and short-range feature dependencies for 3D medical image segmentation. Zhou et al.\cite{zhou2021nnformer} present nnFormer, a method derived from the Swin-UNet\cite{cao2022swin} architecture. Wang et al.\cite{wang2021transbts} proposed TransBTS which uses a regular convolutional encoder-decoder architecture and a Transformer layer as the bottleneck.  These methods suffers from considerable model and computational complexity.

\begin{figure*}[ht]
    \centering
    \begin{subfigure}{10cm}
        \includegraphics[width=\textwidth, height=8cm]{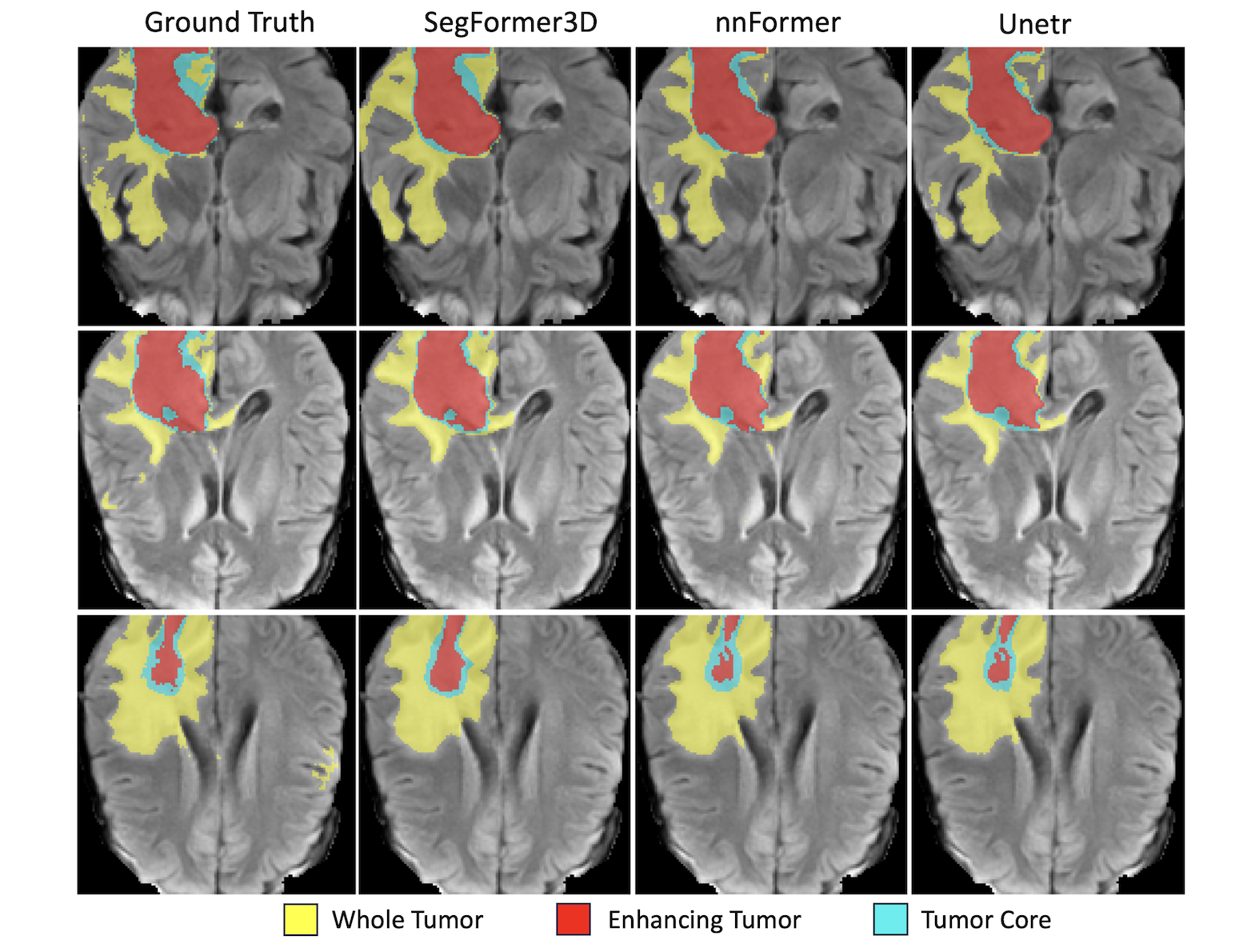}
    \captionsetup{labelformat=empty} 
    \caption{{Fig. 2: Qualitative results on BRaTs. Each row is a separate frame in the MRI sequence while each column is 3D volumetric image segmentation solution. We qualitatively demonstrate highly accurate segmentation performance to SOTA methods while maintaining a lightweight and efficient architecture. }}
    \label{fig: brats}
    \end{subfigure}
    \bigskip
    \begin{subfigure}{8cm}
        \centering
        \renewcommand\tabularxcolumn[1]{m{#1}}
    \begin{adjustbox}{width=8cm,center}
   \begin{tabular}{l|c|c|c|c|c}
    \hline
    Methods & Params & Avg $\%$ $\uparrow$ & \begin{tabular}{@{}c@{}}Whole \\ \vspace{2ex} Tumor $\uparrow$ \end{tabular} & \begin{tabular}{@{}c@{}}Enhancing \\ \vspace{2ex} Tumor $\uparrow$ \end{tabular} & \begin{tabular}{@{}c@{}}Tumor \\ \vspace{2ex} Core $\uparrow$\end{tabular} \\
    \hline
    nnFormer\cite{zhou2021nnformer} & 150.5  & 86.4 & 91.3 & 81.8 & 86.0    \\
    \textbf{Ours} & \textbf{4.5}  & \textbf{82.1} & \textbf{89.9} & \textbf{74.2} & \textbf{82.2}   \\
    UNETR\cite{hatamizadeh2022unetr} & 92.49  & 71.1 & 78.9 & 58.5 & 76.1   \\
    TransBTS\cite{wang2021transbts} & --  & 69.6 & 77.9 & 57.4 & 73.5    \\
    CoTr\cite{Xie2021CoTrEB} & 41.9  & 68.3 & 74.6 & 55.7 & 74.8   \\
    CoTr w/o CNN Encoder\cite{Xie2021CoTrEB} & --  & 64.4 & 71.2 & 52.3 & 69.8  \\
    TransUNet\cite{chen2021transunet} & 96.07  & 64.4 & 70.6 & 54.2 & 68.4    \\
    SETR MLA\cite{Zheng2020RethinkingSS} & 310.5  & 63.9 & 69.8 & 55.4 & 66.5  \\
    SETR PUP\cite{Zheng2020RethinkingSS} & 318.31  & 63.8 & 69.6 & 54.9 & 67.0   \\
    SETR NUP\cite{Zheng2020RethinkingSS} & 305.67  & 63.7 & 69.7 & 54.4 & 66.9   \\
    \hline
  \end{tabular}
    \label{tab: bratstable}
    \end{adjustbox}
    \captionsetup{labelformat=empty} 
        \caption{Table 2: BRaTs comparison table ranked based on average performance across all classes. Segformer3D is highly competitive out performing well established solutions across all categories.}
    \end{subfigure}
    \label{my label}
\end{figure*}

\section{Method}

The adoption of Transformers has greatly improved the performance of volumetric medical image segmentation. However, current high-performing architectures prioritize over-parameterization for model performance, sacrificing efficiency. To demonstrate the benefits of lightweight and efficient Transformers without compromising on performance, we introduce Segformer3D. With \textbf{4.5 million} parameters and \textbf{17 GFLOPS} we show a reduction of $\textbf{34}\times$ and $\textbf{13}\times$ in parameter count and complexity showcasing the significance of the proposed architecture in 3D medical image segmentation \ref{tab:mytable_transposed}.

\begin{table}[h]
  \centering
  \begin{tabular}{l|c|c}
    \toprule
    Architecture & Params & GFLOPs \\
    \midrule
    nnFormer\cite{zhou2021nnformer} & 150.5 & 213.4 \\
    TransUnet\cite{chen2021transunet} & 96.07 & 88.91 \\
    UNETR\cite{hatamizadeh2022unetr} & 92.49 & 75.76 \\
    SwinUNETR\cite{hatamizadeh2021swin} & 62.83 & 384.2 \\
    \textbf{Segformer3D (ours)} & \textbf{4.51} & \textbf{17.5} \\
    \bottomrule
  \end{tabular}
  \caption{Segformer3D vs SOTA in Size (M), and complexity. Segformer3D showcases a significant reduction in parameters and computational complexity without sacrificing on performance.}
  \label{tab:mytable_transposed}
\end{table}


\begin{figure*}[ht]
    \centering
    \begin{subfigure}{10cm}
        \includegraphics[width=\textwidth, height=8cm]{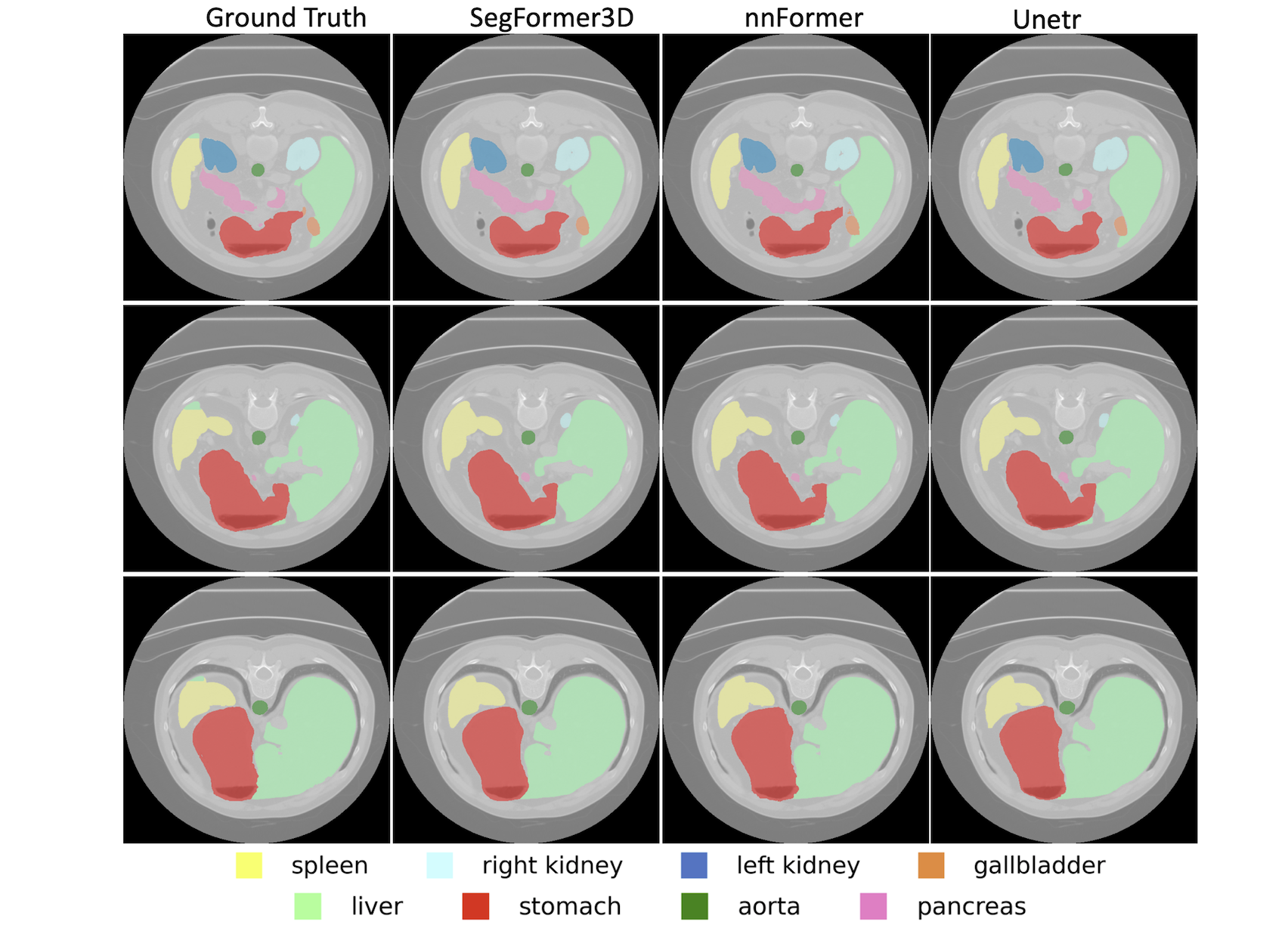}
    \captionsetup{labelformat=empty} 
    \caption{{Fig. 3: Qualitative results on Synapse. Each row is a separate frame in the CT sequence while each column is different 3D volumetric image segmentation solution. Each organ mask is highlighted with a unique color code. We qualitatively demonstrate highly accurate segmentation performance compared to well estabilied SOTA methods while maintaining a lightweight design.}}
    \label{fig: synapse}
    \end{subfigure}
    \bigskip
    \begin{subfigure}{10cm}
        \centering
        \renewcommand\tabularxcolumn[1]{m{#1}}
    \begin{adjustbox}{width=10cm,center}
  \begin{tabular}{l|c|c|c|c|c|c|c|c|c|c}
    \hline
    Methods & Params &  Avg $\%$ $\uparrow$ & AOR & LIV & LKID & RKID & GAL& PAN & SPL & STO                                                     \\
    \hline
    nnFormer\cite{zhou2021nnformer} & 150.5   & 86.57 & 92.04 & 96.84  & 86.57 & 86.25 & 70.17 & 83.35 & 90.51 & 86.83              \\
    \textbf{Ours} & \textbf{4.5}  & \textbf{82.15}  & \textbf{90.43} & \textbf{95.68} & \textbf{86.53} & \textbf{86.13} & \textbf{55.26}& \textbf{73.06} & \textbf{89.02} & \textbf{81.12} \\ 
    MISSFormer\cite{MISSFormer} & --  & 81.96 & 86.99 & 94.41 & 85.21 & 82.00 & 68.65  & 65.67 & 91.92 & 80.81                     \\
    UNETR\cite{hatamizadeh2022unetr} & 92.49   &  79.56 & 89.99 & 94.46 & 85.66 & 84.80 & 60.56  & 59.25 & 87.81 & 73.99           \\
    SwinUNet\cite{cao2022swin} & --   & 79.13 & 85.47 & 94.29 & 83.28 & 79.61 & 66.53  & 56.58 & 90.66 & 76.60                      \\
    LeVit-UNet-384\cite{Xu2021LeViTUNetMF} & 52.17   & 78.53 & 87.33 & 93.11 & 84.61 & 80.25 & 62.23  & 59.07 & 88.86 & 72.76      \\
    TransClaw U-Net\cite{Chang2021TransclawUC} & --  & 78.09 & 85.87 & 94.28 & 84.83 & 79.36 & 61.38  & 57.65 & 87.74 & 73.55      \\
    TransUNet\cite{chen2021transunet} & 96.07  & 77.48 & 87.23 & 94.08 & 81.87 & 77.02 & 63.16  & 55.86 & 85.08 & 75.62            \\
    R50-ViT+CUP\cite{chen2021transunet}  & 86.00  & 71.29 & 73.73 & 91.51 & 75.80 & 72.20 & 55.13  & 45.99 & 81.99 & 73.95         \\
    ViT+CUP\cite{chen2021transunet}  & 86.00  & 67.86 & 70.19 & 91.32 & 74.70 & 67.40 & 45.10  & 42.00 & 81.75 & 70.44             \\
    \hline
  \end{tabular}
    \end{adjustbox}
    \captionsetup{labelformat=empty} 
        \caption{Table 3: Synapse comparisons ranked based on average performance across classes. Segformer3D is highly competitive, outperforming well-established solutions and second to only nnformer with 34x parameters. }
    \end{subfigure}
\end{figure*}

\paragraph{\bf Encoder:} Using 3D medical images within the Transformer framework results in long sequence lengths which increases the computational complexity of the model. For example, a standard 3D MRI volume with dimensions of 128³ results in a sequence length of 32,768, whereas a typical 2D RGB image with dimensions of 256² yields a sequence length of 256. Our hierarchical Transformer incorporates three key elements to improve computational efficiency and reduce the total parameter count while maintaining the SOTA level performance. First, we incorporate overlapped patch merging to overcome neighborhood information loss during the voxel generation process. This technique, in contrast to the patching mechanism seen in ViT\cite{dosovitskiy2020image}, allows the model to better understand the transition points between the voxels and has been shown to improve overall segmentation precision\cite{xie2021segformer}. Next, to address the sequence length bottleneck without compromising performance, we integrate an efficient self-attention mechanism \cite{wang2021pyramid}. This approach enables the model to capture long-range dependencies more effectively, promoting improved scalability and performance. Traditional self-attention takes a sequence of vectors of shape [Batch, Sequence, Features] as input and generates 3 unique projections, Query, Key and Value vectors. Once generated, the attention scores are computed as $(Q, K, V) = \text{Softmax}\left( \frac{QK^T}{\sqrt{d_{\text{head}}}} \right) V$. Due to the operation $QK^T$, the computational complexity of the original segmentation process is $\bigO(n^2)$. Although this complexity can be overlooked with 2D images, with long 3D sequences, it proves to be a challenge for efficient architecture design. Efficient attention introduced in \cite{xie2021segformer, wang2021pyramid}.
\begin{eqnarray}
    \hat{K} &=& \text{Reshape}(\frac{N}{R}, C \cdot R)(K), \nonumber\\
    K &=& \text{Linear}(C \cdot R, C)(\hat{K}) \nonumber,
\end{eqnarray}
 significantly reduces the computational complexity generated by 3D volumetric tensors from $\bigO(n^2)$ to $\bigO(n^2/r)$. We set the reduction parameter $r$ to $4\times$, $2\times$, $1\times$, $1\times$ in the four stages of the encoder. 

Finally, our approach addresses the challenge of resizing volumetric imaging and its relation to fixed positional encoding in ViTs by adopting the mix-ffn module \cite{xie2021segformer}. This module enables automatic learning of positional cues, eliminating the need for fixed encoding, ensuring superior scalability and performance. 


\paragraph{\bf Decoder:} The decoding stage plays a pivotal role in medical image segmentation based on the encoder-decoder design widely adopted in the UNET based architectures\cite{hatamizadeh2021swin, hatamizadeh2022unetr}. This framework is used in both CNN-based and Transformer-based encoders. In the context of 3D medical images, where successive 3D convolutions are often necessary for effective decoding, we instead demonstrate that the integration of linear layers is a highly effective decoding strategy for medical image segmentation. Our approach simplifies the decoding process, ensuring efficient and consistent decoding of volumetric features across diverse datasets without over-parameterization. The simple decoder process is:
\vspace{-\baselineskip} 
\begin{eqnarray}
    \text{step 1: }&     F_i &= \text{Linear}(C_i, C)(F_i), \quad \forall i\\
    \text{step 2: }&     \hat{F}_i &= \text{Upsample}\left(W_{4 \times 4}\right)(\hat{F}_i), \quad \forall i\\
    \text{step 3: }&     F &= \text{Linear}(4C, C)(\text{Concat}(\hat{F}_i)), \quad \forall i\\
    \text{step 4: }&     M &= \text{Linear}(C, N_{\text{cls}})(F)
\end{eqnarray}
Similar to the skip connections introduced in UNET, features at each stage are collated, and a fixed dimensional projection is generated. Once all dimensions are standardized, we upsample each feature and perform a concatenation followed by a fusion operation. The fused features are input to a linear projection head (3D 1x1 convolutions) to generate the final segmentation masks. 

\begin{figure*}[ht]
    \centering
    \begin{subfigure}{10cm}
        \includegraphics[width=\textwidth, height=8cm]{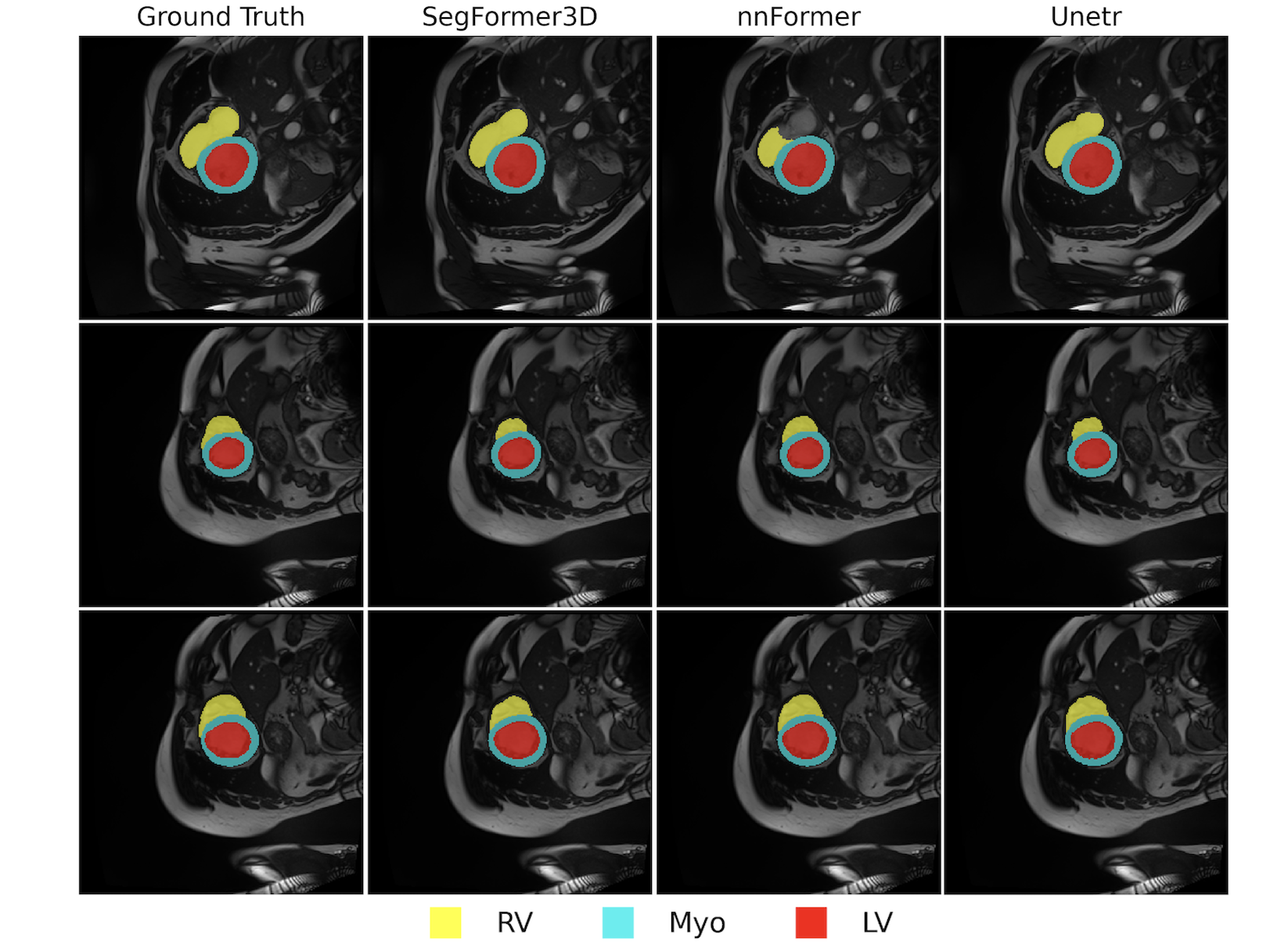}
    \captionsetup{labelformat=empty} 
    \caption{Fig. 4: Qualitative results on ACDC. Each row is a separate frame in the cine-MRI sequence while each column is different 3D volumetric image segmentation solution. We demonstrate highly accurate segmentation results to SOTA methods while maintaining a lightweight and efficient architecture.}
    \label{fig: acdc}
    \end{subfigure}
    \bigskip
    \begin{subfigure}{8cm}
        \centering
    \begin{adjustbox}{width=8cm,center}
    \begin{tabular}{l|c|c|c|c|c}
    \hline
    Methods & Params & Avg $\%$ $\uparrow$ & RV & Myo & LV                                           \\
    \hline
    nnFormer \cite{zhou2021nnformer} & 150.5  & 92.06 & 90.94 & 89.58 & 95.65            \\
    \textbf{Ours} & \textbf{4.5}  & \textbf{90.96} & \textbf{88.50} & \textbf{88.86} & \textbf{95.53} \\
    LeViT-UNet-384 \cite{Xu2021LeViTUNetMF} & 52.17  & 90.32 & 89.55 & 87.64 & 93.76     \\
    SwinUNet \cite{cao2022swin} & --  & 90.00 & 88.55 & 85.62 & 95.83                       \\
    TransUNet \cite{chen2021transunet} & 96.07  & 89.71 & 88.86 & 84.54 & 95.73         \\
    UNETR \cite{hatamizadeh2022unetr} & 92.49 & 88.61 & 85.29 & 86.52 & 94.02           \\
    R50-VIT-CUP \cite{chen2021transunet} & 86.00  & 87.57 & 86.07 & 81.88 & 94.75           \\
    VIT-CUP \cite{chen2021transunet} & 86.00  & 81.45 & 81.46 & 70.71 & 92.18               \\
    \hline
    \end{tabular}
    \label{tab: acdctable}
    \end{adjustbox}
    \captionsetup{labelformat=empty} 
        \caption{Table 4: ACDC comparison ranked based on average performance across  classes. Segformer3D is highly competitive outperforming well established solutions and is within 1\% of SOTA with 150 million parameters.}
    \end{subfigure}
\end{figure*}
\section{Experimental Results}

Adhering to the SOTA architecture for 3D volumetric segmentation, we utilize the same datasets and evaluation methods to ensure a fair and consistent comparison across all architectures. We train and evaluate the proposed model on three widely used datasets without the use of external data for pretraining purposes. These datasets are Brain Tumor Segmentation (BraTS) \cite{menze2014multimodal}, Synapse Multi-Organ Segmentation (Synapse) \cite{landman2015miccai}, and finally the Automatic Cardiac Diagnosis (ACDC) \cite{bernard2018deep} datasets.

All experiments, including training, real-time augmentation, and inference, were performed on a single Nvidia RTX 3090 GPU using PyTorch. Each model is trained with the same learning rate policy, which includes a learning rate warm-up stage, where we linearly increase the learning rate from $4e-6$ to $4e-4$, which is followed by a PolyLR decay strategy. The widely adopted AdamW optimizer \cite{Loshchilov2017DecoupledWD} was used with a learning rate of $3e-5$. For the loss function, an equally weighted Dice-Cross Entropy Loss combination was adopted to combine the benefits of each loss function in the optimization process, improving convergence. We set the batch size to 4 and train each model for 1000 epochs, similar to the SOTA architecture. Additionally, all experiments are performed without the use of complicated pretraining efforts to showcase the proposed architectures performance on real-world medical datasets without additional data. 

\subsection{Results on Brain Tumor Segmentation (BraTs)}
BraTs \cite{menze2014multimodal} is a dataset for medical image segmentation from MRI scans. The dataset contains 484 MRI images with four modalities, FLAIR, T1w, T1gd, and T2w. Data were collected from 19 institutions with ground-truth labels for three types of tumor subregions: edema (ED), enhancing tumor (ET) and nonenhancing tumor (NET). Following the same data preparation, augmentation and reporting strategies in major publications including nnFormer \cite{zhou2021nnformer} we report our results on whole tumor (WT), enhancing tumor (ET) and tumor core (TC). In Table \ref{fig: brats} we demonstrate that Segformer3D stands out as a strong competitive lightweight segmentation architecture against significantly larger and widely adopted CNN and Transformer architectures while maintaining 4.5 million parameters and 17.5 GFLOP computational complexity. This demonstrates the representation learning capability of the efficient self attention module over traditional ViT architectures that analyzes the whole sequence of patches without compression. Finally, we showcase highly competitive qualitative results of the proposed architecture in Figure \ref{fig: brats} showcasing superior performance against well established architectures.

\subsection{Results on Multi-Organ CT Segmentation (Synapse)}

The Synapse dataset \cite{landman2015miccai} provides 30 annotated CT images. We generate our results after data processing, training, and data splits defined in \cite{zhou2021nnformer}. With a diverse set of annotations that cover multiple organs, such as spleen, pancreas, gallbladder, and others, the synapse dataset is a complex multi-class segmentation challenge. The quantitative results in Table \ref{fig: synapse} show that Segformer3D is ranked second only to the nnFormer \cite{zhou2021nnformer} architecture with 150 million parameters. Additionally, we showcase qualitative performance results in Figure \ref{fig: synapse} where we compare highly accurate organ segmentation masks with current SOTA architectures validating the visual consistency of the proposed approach. Finally, compared to widely used architectures \cite{hatamizadeh2022unetr, MISSFormer, hatamizadeh2021swin}, Segformer3D generates competitive results with only 4.5M parameters and demonstrates that over parameterization does not lead to large performance gains, especially in data constrained situations. 

\subsection{Results on Automated cardiac diagnosis (ACDC)}

ACDC\cite{bernard2018deep} is a dataset of 100 patients used for 3D volumetric segmentation of the left (LV) and right (RV) cardiac ventricles and the myocardium (Myo)\cite{bernard2018deep}. To maintain a one-to-one comparison with published research, we follow the same training and inference pipelines specified in \cite{zhou2021nnformer} and measure segmentation accuracy using the Dice metric. Table \ref{fig: acdc} quantitatively demonstrates the proposed architecture is highly competitive against large and highly complex solutions. The proposed model is within ~1\% margin of the SOTA performance with models on average $34\times$ higher in parameter count and $13\times$ higher in computational complexity. Comparisons of the qualitative results are visualized in Figure \ref{fig: acdc} showcasing highly competitive performance without the need for large scale pretraining on small datasets.

\subsection{Conclusion}

Architectures such as UNETR, TransUNet and nnFormer have revolutionized 3D volumetric medical image segmentation using the ViT framework. This paradigm shift has notably enhanced the model's contextual understanding capabilities compared to its conventional pure Convolutional Neural Network (CNN) counterparts. However, this improvement has come at the cost of a substantial increase in parameter count and model complexity, attributed to the intricate nature of the self-attention module. In addition to model size and complexities, large models prevent medical researchers with limited access to large scale compute resources from effectively training and integrating these models into their workflows. Furthermore, larger models can introduce challenges to model generalization, and convergence, especially in scenarios with limited datasets commonly seen in medical imaging. To overcome these limitations without sacrificing on performance we introduce Segformer3D a lightweight architecture that is $34\times$ smaller in parameters and $13\times$ less in parameters and computational complexity respectively over the state-of-the-art (SOTA) architectures. We benchmark our solution to current SOTA solutions as well as other highly cited works and we showcase that lightweight and efficient architectures can help significantly improve performance over much larger models without additional pretraining and with minimal computational resources. Finally, we assert that directing research efforts toward the development of high-performance lightweight architectures, particularly in domains with tangible real-world advantages like medical imaging, not only broadens accessibility but also promotes practical applications of such architectures in real-world scenarios.

{\small
\bibliographystyle{ieeenat_fullname}
\bibliography{references}

\begin{thebibliography}{33}
\providecommand{\natexlab}[1]{#1}
\providecommand{\url}[1]{\texttt{#1}}
\expandafter\ifx\csname urlstyle\endcsname\relax
  \providecommand{\doi}[1]{doi: #1}\else
  \providecommand{\doi}{doi: \begingroup \urlstyle{rm}\Url}\fi

\bibitem[Bernard et~al.(2018)Bernard, Lalande, Zotti, Cervenansky, Yang, Heng, Cetin, Lekadir, Camara, Ballester, et~al.]{bernard2018deep}
Olivier Bernard, Alain Lalande, Clement Zotti, Frederick Cervenansky, Xin Yang, Pheng-Ann Heng, Irem Cetin, Karim Lekadir, Oscar Camara, Miguel Angel~Gonzalez Ballester, et~al.
\newblock Deep learning techniques for automatic mri cardiac multi-structures segmentation and diagnosis: is the problem solved?
\newblock \emph{IEEE transactions on medical imaging}, 37\penalty0 (11), 2018.

\bibitem[Cai et~al.(2020)Cai, Tian, Lui, Zeng, Wu, and Chen]{cai2020dense}
Sijing Cai, Yunxian Tian, Harvey Lui, Haishan Zeng, Yi Wu, and Guannan Chen.
\newblock Dense-unet: a novel multiphoton in vivo cellular image segmentation model based on a convolutional neural network.
\newblock \emph{Quantitative imaging in medicine and surgery}, 10\penalty0 (6):\penalty0 1275, 2020.

\bibitem[Cao et~al.(2022)Cao, Wang, Chen, Jiang, Zhang, Tian, and Wang]{cao2022swin}
Hu Cao, Yueyue Wang, Joy Chen, Dongsheng Jiang, Xiaopeng Zhang, Qi Tian, and Manning Wang.
\newblock Swin-unet: Unet-like pure transformer for medical image segmentation.
\newblock In \emph{European conference on computer vision}. Springer, 2022.

\bibitem[Chang et~al.(2021)Chang, Hu, Zhai, and Zhang]{Chang2021TransclawUC}
Yao Chang, Menghan Hu, Guangtao Zhai, and Xiao-Ping Zhang.
\newblock Transclaw u-net: Claw u-net with transformers for medical image segmentation.
\newblock \emph{2022 5th International Conference on Information Communication and Signal Processing (ICICSP)}, 2021.

\bibitem[Chen et~al.(2021)Chen, Lu, Yu, Luo, Adeli, Wang, Lu, Yuille, and Zhou]{chen2021transunet}
Jieneng Chen, Yongyi Lu, Qihang Yu, Xiangde Luo, Ehsan Adeli, Yan Wang, Le Lu, Alan~L Yuille, and Yuyin Zhou.
\newblock Transunet: Transformers make strong encoders for medical image segmentation.
\newblock \emph{arXiv:2102.04306}, 2021.

\bibitem[{\c{C}}i{\c{c}}ek et~al.(2016){\c{C}}i{\c{c}}ek, Abdulkadir, Lienkamp, Brox, and Ronneberger]{cciccek20163d}
{\"O}zg{\"u}n {\c{C}}i{\c{c}}ek, Ahmed Abdulkadir, Soeren~S Lienkamp, Thomas Brox, and Olaf Ronneberger.
\newblock 3d u-net: learning dense volumetric segmentation from sparse annotation.
\newblock In \emph{Medical Image Computing and Computer-Assisted Intervention--MICCAI 2016: 19th International Conference, Athens, Greece, October 17-21, 2016, Proceedings, Part II 19}, pages 424--432. Springer, 2016.

\bibitem[Dosovitskiy et~al.(2020)Dosovitskiy, Beyer, Kolesnikov, Weissenborn, Zhai, Unterthiner, Dehghani, Minderer, Heigold, Gelly, et~al.]{dosovitskiy2020image}
Alexey Dosovitskiy, Lucas Beyer, Alexander Kolesnikov, Dirk Weissenborn, Xiaohua Zhai, Thomas Unterthiner, Mostafa Dehghani, Matthias Minderer, Georg Heigold, Sylvain Gelly, et~al.
\newblock An image is worth 16x16 words: Transformers for image recognition at scale.
\newblock \emph{arXiv:2010.11929}, 2020.

\bibitem[Dou et~al.(2016)Dou, Chen, Jin, Yu, Qin, and Heng]{dou20163d}
Qi Dou, Hao Chen, Yueming Jin, Lequan Yu, Jing Qin, and Pheng-Ann Heng.
\newblock 3d deeply supervised network for automatic liver segmentation from ct volumes.
\newblock In \emph{Medical Image Computing and Computer-Assisted Intervention--MICCAI 2016: 19th International Conference, Athens, Greece, October 17-21, 2016, Proceedings, Part II 19}. Springer, 2016.

\bibitem[Gibson et~al.(2018)Gibson, Giganti, Hu, Bonmati, Bandula, Gurusamy, Davidson, Pereira, Clarkson, and Barratt]{gibson2018automatic}
Eli Gibson, Francesco Giganti, Yipeng Hu, Ester Bonmati, Steve Bandula, Kurinchi Gurusamy, Brian Davidson, Stephen~P Pereira, Matthew~J Clarkson, and Dean~C Barratt.
\newblock Automatic multi-organ segmentation on abdominal ct with dense v-networks.
\newblock \emph{IEEE transactions on medical imaging}, 37\penalty0 (8), 2018.

\bibitem[Hatamizadeh et~al.(2021)Hatamizadeh, Nath, Tang, Yang, Roth, and Xu]{hatamizadeh2021swin}
Ali Hatamizadeh, Vishwesh Nath, Yucheng Tang, Dong Yang, Holger~R Roth, and Daguang Xu.
\newblock Swin unetr: Swin transformers for semantic segmentation of brain tumors in mri images.
\newblock In \emph{International MICCAI Brainlesion Workshop}. Springer, 2021.

\bibitem[Hatamizadeh et~al.(2022)Hatamizadeh, Tang, Nath, Yang, Myronenko, Landman, Roth, and Xu]{hatamizadeh2022unetr}
Ali Hatamizadeh, Yucheng Tang, Vishwesh Nath, Dong Yang, Andriy Myronenko, Bennett Landman, Holger~R Roth, and Daguang Xu.
\newblock Unetr: Transformers for 3d medical image segmentation.
\newblock In \emph{Proceedings of the IEEE/CVF winter conference on applications of computer vision}, 2022.

\bibitem[Huang et~al.(2023)Huang, Deng, Li, Yuan, and Fu]{MISSFormer}
Xiaohong Huang, Zhifang Deng, Dandan Li, Xueguang Yuan, and Ying Fu.
\newblock Missformer: An effective transformer for 2d medical image segmentation.
\newblock \emph{IEEE Transactions on Medical Imaging}, 42\penalty0 (5), 2023.

\bibitem[Isensee et~al.(2018)Isensee, Petersen, Klein, Zimmerer, Jaeger, Kohl, Wasserthal, Koehler, Norajitra, Wirkert, and Maier-Hein]{isensee2018nnunet}
Fabian Isensee, Jens Petersen, Andre Klein, David Zimmerer, Paul~F. Jaeger, Simon Kohl, Jakob Wasserthal, Gregor Koehler, Tobias Norajitra, Sebastian Wirkert, and Klaus~H. Maier-Hein.
\newblock nnu-net: Self-adapting framework for u-net-based medical image segmentation, 2018.

\bibitem[Islam et~al.(2020)Islam, Jia, and Bruce]{islam2020much}
Md~Amirul Islam, Sen Jia, and Neil~DB Bruce.
\newblock How much position information do convolutional neural networks encode?
\newblock \emph{arXiv:2001.08248}, 2020.

\bibitem[Landman et~al.(2015)Landman, Xu, Igelsias, Styner, Langerak, and Klein]{landman2015miccai}
Bennett Landman, Zhoubing Xu, J Igelsias, Martin Styner, T Langerak, and Arno Klein.
\newblock Miccai multi-atlas labeling beyond the cranial vault--workshop and challenge.
\newblock In \emph{Proc. MICCAI Multi-Atlas Labeling Beyond Cranial Vault—Workshop Challenge}, 2015.

\bibitem[Li et~al.(2020)Li, Pan, Zhu, and Qin]{li2020pgdunet}
Ziqiang Li, Hong Pan, Yaping Zhu, and A~Kai Qin.
\newblock Pgd-unet: A position-guided deformable network for simultaneous segmentation of organs and tumors.
\newblock In \emph{2020 International Joint Conference on Neural Networks (IJCNN)}. IEEE, 2020.

\bibitem[Liu et~al.(2023)Liu, Peng, Zheng, Yang, Hu, and Yuan]{liu2023efficientvit}
Xinyu Liu, Houwen Peng, Ningxin Zheng, Yuqing Yang, Han Hu, and Yixuan Yuan.
\newblock Efficientvit: Memory efficient vision transformer with cascaded group attention.
\newblock In \emph{Proceedings of the IEEE/CVF Conference on Computer Vision and Pattern Recognition}, pages 14420--14430, 2023.

\bibitem[Long et~al.(2015)Long, Shelhamer, and Darrell]{long2015fully}
Jonathan Long, Evan Shelhamer, and Trevor Darrell.
\newblock Fully convolutional networks for semantic segmentation.
\newblock In \emph{Proceedings of the IEEE conference on computer vision and pattern recognition}, pages 3431--3440, 2015.

\bibitem[Loshchilov and Hutter(2017)]{Loshchilov2017DecoupledWD}
Ilya Loshchilov and Frank Hutter.
\newblock Decoupled weight decay regularization.
\newblock In \emph{International Conference on Learning Representations}, 2017.

\bibitem[Menze et~al.(2014)Menze, Jakab, Bauer, Kalpathy-Cramer, Farahani, Kirby, Burren, Porz, Slotboom, Wiest, et~al.]{menze2014multimodal}
Bjoern~H Menze, Andras Jakab, Stefan Bauer, Jayashree Kalpathy-Cramer, Keyvan Farahani, Justin Kirby, Yuliya Burren, Nicole Porz, Johannes Slotboom, Roland Wiest, et~al.
\newblock The multimodal brain tumor image segmentation benchmark (brats).
\newblock \emph{IEEE transactions on medical imaging}, 34\penalty0 (10), 2014.

\bibitem[Milletari et~al.(2016)Milletari, Navab, and Ahmadi]{milletari2016vnet}
Fausto Milletari, Nassir Navab, and Seyed-Ahmad Ahmadi.
\newblock V-net: Fully convolutional neural networks for volumetric medical image segmentation.
\newblock In \emph{2016 fourth international conference on 3D vision (3DV)}, pages 565--571. Ieee, 2016.

\bibitem[Monsefi et~al.(2024)Monsefi, Karisani, Zhou, Choi, Doble, Ji, Parthasarathy, and Ramnath]{monsefi2024masked}
Amin~Karimi Monsefi, Payam Karisani, Mengxi Zhou, Stacey Choi, Nathan Doble, Heng Ji, Srinivasan Parthasarathy, and Rajiv Ramnath.
\newblock Masked logonet: Fast and accurate 3d image analysis for medical domain.
\newblock \emph{arXiv preprint arXiv:2402.06190}, 2024.

\bibitem[Ronneberger et~al.(2015)Ronneberger, Fischer, and Brox]{ronneberger2015u}
Olaf Ronneberger, Philipp Fischer, and Thomas Brox.
\newblock U-net: Convolutional networks for biomedical image segmentation.
\newblock In \emph{Medical Image Computing and Computer-Assisted Intervention--MICCAI 2015: 18th International Conference, Munich, Germany, October 5-9, 2015, Proceedings, Part III 18}. Springer, 2015.

\bibitem[Roth et~al.(2017)Roth, Oda, Hayashi, Oda, Shimizu, Fujiwara, Misawa, and Mori]{roth2017hierarchical}
Holger~R Roth, Hirohisa Oda, Yuichiro Hayashi, Masahiro Oda, Natsuki Shimizu, Michitaka Fujiwara, Kazunari Misawa, and Kensaku Mori.
\newblock Hierarchical 3d fully convolutional networks for multi-organ segmentation.
\newblock \emph{arXiv:1704.06382}, 2017.

\bibitem[Wang et~al.(2021{\natexlab{a}})Wang, Chen, Ding, Yu, Zha, and Li]{wang2021transbts}
Wenxuan Wang, Chen Chen, Meng Ding, Hong Yu, Sen Zha, and Jiangyun Li.
\newblock Transbts: Multimodal brain tumor segmentation using transformer.
\newblock In \emph{Medical Image Computing and Computer Assisted Intervention--MICCAI 2021: 24th International Conference, Strasbourg, France, September 27--October 1, 2021, Proceedings, Part I 24}. Springer, 2021{\natexlab{a}}.

\bibitem[Wang et~al.(2021{\natexlab{b}})Wang, Xie, Li, Fan, Song, Liang, Lu, Luo, and Shao]{wang2021pyramid}
Wenhai Wang, Enze Xie, Xiang Li, Deng-Ping Fan, Kaitao Song, Ding Liang, Tong Lu, Ping Luo, and Ling Shao.
\newblock Pyramid vision transformer: A versatile backbone for dense prediction without convolutions.
\newblock In \emph{Proceedings of the IEEE/CVF international conference on computer vision}, 2021{\natexlab{b}}.

\bibitem[Xie et~al.(2021{\natexlab{a}})Xie, Wang, Yu, Anandkumar, Alvarez, and Luo]{xie2021segformer}
Enze Xie, Wenhai Wang, Zhiding Yu, Anima Anandkumar, Jose~M Alvarez, and Ping Luo.
\newblock Segformer: Simple and efficient design for semantic segmentation with transformers.
\newblock \emph{Advances in Neural Information Processing}, 34, 2021{\natexlab{a}}.

\bibitem[Xie et~al.(2021{\natexlab{b}})Xie, Zhang, Shen, and Xia]{Xie2021CoTrEB}
Yutong Xie, Jianpeng Zhang, Chunhua Shen, and Yong Xia.
\newblock Cotr: Efficiently bridging cnn and transformer for 3d medical image segmentation.
\newblock \emph{ArXiv}, abs/2103.03024, 2021{\natexlab{b}}.

\bibitem[Xu et~al.(2021)Xu, Wu, Zhang, and He]{Xu2021LeViTUNetMF}
Guoping Xu, Xingrong Wu, Xuan Zhang, and Xinwei He.
\newblock Levit-unet: Make faster encoders with transformer for medical image segmentation.
\newblock \emph{ArXiv}, abs/2107.08623, 2021.

\bibitem[Zhang et~al.(2021)Zhang, Liu, and Hu]{zhang2021transfuse}
Yundong Zhang, Huiye Liu, and Qiang Hu.
\newblock Transfuse: Fusing transformers and cnns for medical image segmentation.
\newblock In \emph{Medical Image Computing and Computer Assisted Intervention--MICCAI 2021}. Springer, 2021.

\bibitem[Zheng et~al.(2020)Zheng, Lu, Zhao, Zhu, Luo, Wang, Fu, Feng, Xiang, Torr, and Zhang]{Zheng2020RethinkingSS}
Sixiao Zheng, Jiachen Lu, Hengshuang Zhao, Xiatian Zhu, Zekun Luo, Yabiao Wang, Yanwei Fu, Jianfeng Feng, Tao Xiang, Philip H.~S. Torr, and Li Zhang.
\newblock Rethinking semantic segmentation from a sequence-to-sequence perspective with transformers.
\newblock \emph{2021 IEEE/CVF Conference on Computer Vision and Pattern Recognition (CVPR)}, 2020.

\bibitem[Zhou et~al.(2021)Zhou, Guo, Zhang, Yu, Wang, and Yu]{zhou2021nnformer}
Hong-Yu Zhou, Jiansen Guo, Yinghao Zhang, Lequan Yu, Liansheng Wang, and Yizhou Yu.
\newblock nnformer: Interleaved transformer for volumetric segmentation.
\newblock \emph{arXiv:2109.03201}, 2021.

\bibitem[Zhu et~al.(2017)Zhu, Du, Turkbey, Choyke, and Yan]{zhu2017deeply}
Qikui Zhu, Bo Du, Baris Turkbey, Peter~L Choyke, and Pingkun Yan.
\newblock Deeply-supervised cnn for prostate segmentation.
\newblock In \emph{2017 international joint conference on neural networks (IJCNN)}. IEEE, 2017.

\end{thebibliography}
}

\end{document}